\documentclass[conference]{IEEEtran}
\IEEEoverridecommandlockouts
\usepackage{cite}
\usepackage{amsmath,amssymb,amsfonts}
\usepackage{algorithmic}
\usepackage{graphicx}
\usepackage{textcomp}
\usepackage{xcolor}
\usepackage{flushend} 
\usepackage{multirow}
\usepackage{tikz}
\usepackage{pgfplots}
\pgfplotsset{compat=1.17}
\usepackage{hyperref}
\usepackage[noabbrev]{cleveref}

\begin{document}

\title{Video object detection for privacy-preserving patient monitoring in intensive care}


\author{
    \IEEEauthorblockN{Raphael Emberger\IEEEauthorrefmark{1},  
        Jens Michael Boss\IEEEauthorrefmark{2},  
        Daniel Baumann\IEEEauthorrefmark{2},
        Marko Seric\IEEEauthorrefmark{2},
        Shufan Huo\IEEEauthorrefmark{2}\IEEEauthorrefmark{3},  
        Lukas Tuggener\IEEEauthorrefmark{1},\\  
        Emanuela Keller\IEEEauthorrefmark{2},  
        Thilo Stadelmann\IEEEauthorrefmark{1}\IEEEauthorrefmark{4}  
    }
    \IEEEauthorblockA{\{embe, tugg, stdm\}@zhaw.ch, \{firstnames.surname\}@usz.ch}
    \IEEEauthorblockA{\IEEEauthorrefmark{1}Centre for Artificial Intelligence, ZHAW School of Engineering, Winterthur, Switzerland}
    \IEEEauthorblockA{\IEEEauthorrefmark{2}Neurocritical Care Unit, Department of Neurosurgery and Institute of Intensive Care Medicine,\\Clinical Neuroscience Center, University Hospital Zurich and University of Zurich, Switzerland}
    \IEEEauthorblockA{\IEEEauthorrefmark{3}Neurology, Charité - University Medicine Berlin, Berlin, Germany}
    \IEEEauthorblockA{\IEEEauthorrefmark{4}European Centre for Living Technology (ECLT), Ca' Bottacin, Venice, Italy}
}


\maketitle

\begin{abstract}
Patient monitoring in intensive care units, although assisted by biosensors, needs continuous supervision of staff.
To reduce the burden on staff members, IT infrastructures are built to record monitoring data and develop clinical decision support systems.
These systems, however, are vulnerable to artifacts (e.g. muscle movement due to ongoing treatment), which are often indistinguishable from real and potentially dangerous signals.
Video recordings could facilitate the reliable classification of biosignals using object detection (OD) methods to find sources of unwanted artifacts.
Due to privacy restrictions, only blurred videos can be stored, which severely impairs the possibility to detect clinically relevant events such as interventions or changes in patient status with standard OD methods.
Hence, new kinds of approaches are necessary that exploit every kind of available information due to the reduced information content of blurred footage and that are at the same time easily implementable within the IT infrastructure of a normal hospital.
In this paper, we propose a new method for exploiting information in the temporal succession of video frames.
To be efficiently implementable using off-the-shelf object detectors that comply with given hardware constraints, we repurpose the image color channels to account for temporal consistency, leading to an improved detection rate of the object classes.
Our method outperforms a standard YOLOv5 baseline model by +1.7\% mAP@.5 while also training over ten times faster on our proprietary dataset.
We conclude that this approach has shown effectiveness in the preliminary experiments and holds potential for more general video OD in the future.
\end{abstract}


\section{Introduction}
\label{sec:Intro}

The intensive care unit (ICU) is a challenging work environment, which demands high staffing and constant alertness toward emergencies.
Numbers, curves, and alarms from multiple medical devices, although well intended, often cause additional stress.
As a consequence, severe burnout syndrome is present in about $50$\% of critical care physicians and one-third of critical care nurses \cite{Embriaco2007a}, which in turn has been shown to correlate strongly with intent to seek other career opportunities.
This is particularly problematic as the healthcare labor shortage has been exacerbated since the onset of the Covid-19 pandemic in many European countries.

To reduce the burden on healthcare professionals and physicians, clinical decision support systems, and early warning systems promise to assist healthcare professionals in decision-making and outcome prediction. Thus avoiding cognitive overload and consequent treatment errors.
These systems take advantage of the vast number of biosignals generated at high resolution by patient monitors and other medical devices.
In a neurocritical care setting, these biosignals include for example arterial pressure, intracranial pressure, blood, and brain tissue oxygenation, electrocardiography, and electroencephalography recordings.
Despite the crucial role these biosignals play in clinical patient assessment, the clinical implementation of machine learning solutions taking full advantage of them is hindered by artifacts in the signals as well as a lack of context in which the signals were acquired \cite{elezi2020exploiting}.
Artifacts can for example be caused by patient motion or staff interventions.
However, without appropriate contextual knowledge, it is not possible to correctly interpret biosignals and distinguish physiological features from artifacts, even though many domain-specific signal processing techniques have been developed \cite{Islam2021}.

\begin{figure*}[t]
    \centering
    \includegraphics[width=\textwidth,trim={0 0 0 1.1cm},clip]{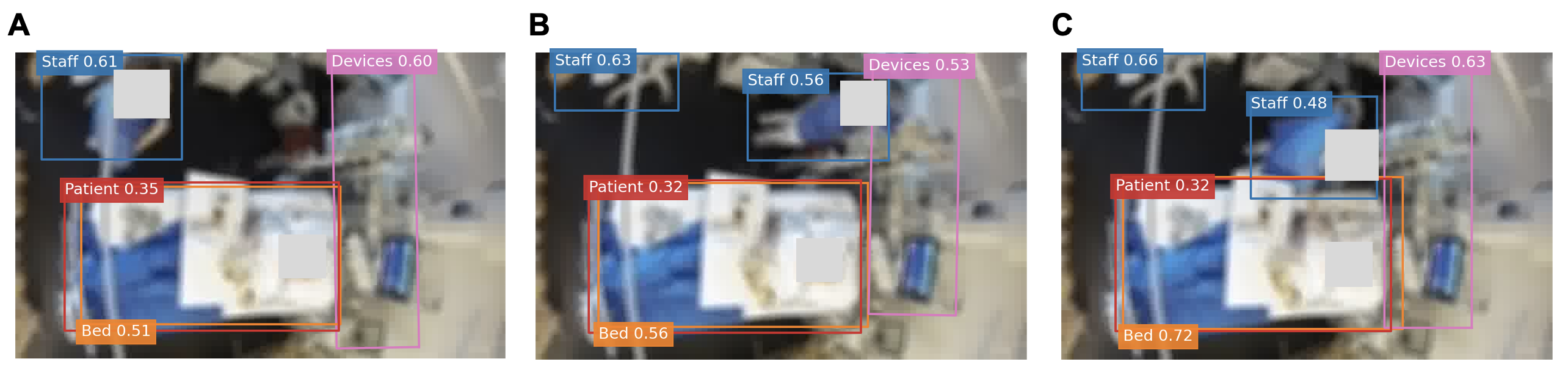}
    \includegraphics[width=\textwidth,trim={0.5cm 0 0 1.1cm},clip]{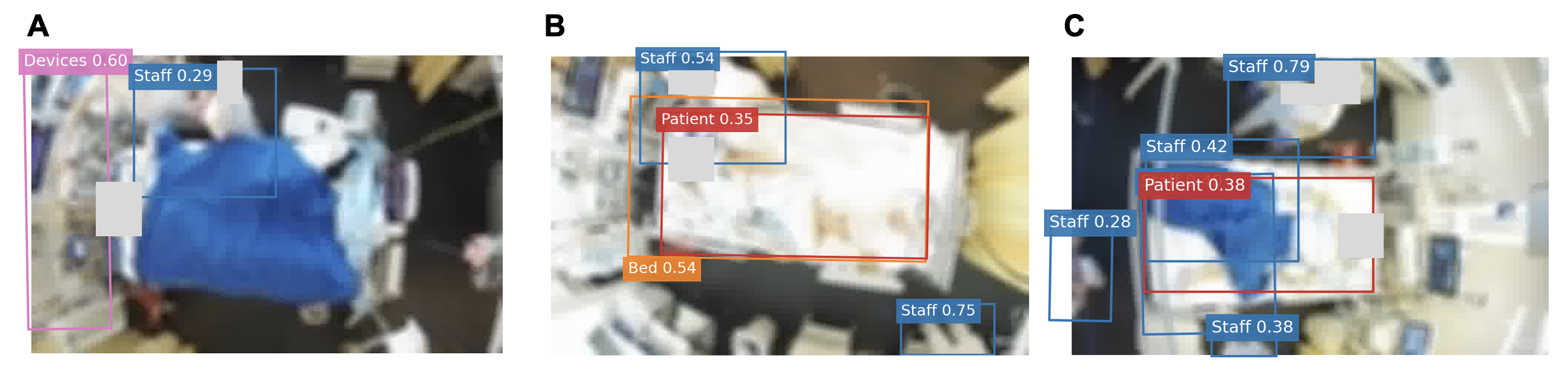}
    \caption{
    Examples from our dataset with overlayed detections of the proposed model.
    \emph{Top row}:
        Successive frames a typical scenario.
        In the first frame (\emph{left}), all objects are correctly detected.
        \emph{Middle} and \emph{right} frames show how the healthcare staff is tracked moving along the bed to a medical device and then to the patient's bedside.
        A white stand at the top left is wrongly identified as staff.
    \emph{Second row}:
        Out-of-context examples where even human observers have difficulty recognizing objects.
        \emph{Left}, the patient is completely covered by a blue blanket, occluding the patient and bed.
        \emph{Middle}, the lighting is comparatively bright, making it difficult for the model to detect the medical devices.
        \emph{Right}, the blue blanket was confused with a staff member's blue jacket.
    }
    \label{fig:dataset-examples}
\end{figure*}

To address this challenge and to gain access to contextual information, we have implemented a camera monitoring system to detect the presence of patients and staff members, thus laying the foundation for more accurate artifact removal approaches.
Ultimately resulting in better decision support systems and thus better patient outcomes, while decreasing the burden on clinical staff by false alarms.
However, the system is subject to the following constraints:
    (a) To respect the privacy of patients, staff, and visitors, video footage can only be stored severely blurred (see Figure \ref{fig:dataset-examples}), removing most of the visual cues to detect relevant objects.
    (b) Also for privacy reasons, the system has to run on-site on hospital hardware resulting in narrow computational constraints.

In this paper, we propose a video OD method to address the preceding challenges, enabling privacy-preserving patient monitoring in clinical practice.
Specifically, our contribution is the extension of a lightweight off-the-shelf still image OD method (that can efficiently run on standard hardware) to learn from the temporal succession of video frames without architectural changes (such that implementation and integration can be performed efficiently).
Experimental evaluation shows +$1.7$\% improvement in mean average precision with $0.5$ IoU overlap while training over ten times faster than the baseline.

\section{Related Work}
\label{sec:RelatedWork}
Camera-based patient motion detection for measuring vital signs and false alarm reduction has been studied in many contexts employing optical flow and artificial neural networks, as well as using 3D cameras~\cite{Muroi2020, Dosso2020, Coronel2021}.
However, even though simple motion quantification approaches showed promising results, they do not provide the same amount of context information as OD methods \cite{jiao2021new}, which yield time-resolved position and class of objects visible in each frame.
OD for patient monitoring has not been studied widely in the literature.
Existing studies have incompatible prerequisites like high-resolution video input \cite{9128812}, or employ off-the-shelf OD methods \cite{electronics9121993} on unblurred data.
The quasi-industry standard for state-of-the-art OD method (also used in \cite{electronics9121993}) is the YOLO family of models \cite{redmon2016you, diwan2022object}.
Of these, the YOLOv5 variant~\cite{Jocher2022} is closest to the application context of this work, as it allows for oriented bounding boxes~\cite{hukaixuan199706272021} and is considered one of the best performing, easy to use and lightweight models.

From a still image of the clips in our dataset, it is hard even for humans to pick out where members of staff are in a frame, due to the high blur and lack of context (cp. \Cref{fig:dataset-examples}).
However, once the same frame is seen in the context of a video, the motions are made visible to viewers and it is easier to identify members of staff.
YOLOV \cite{Ge2021} can leverage chronological video frames stacked together as one sample, exploiting motion information in our dataset.
However, preliminary experiments have shown that this model is not suitable for this application.
Also, domain adaptation techniques to leverage larger pretrained models \cite{sager2022unsupervised} have been trialed, but rejected after preliminary experiments.

\section{A Method for Privacy-Preserving Video OD}
\label{sec:Method}
For the baseline detector, YOLOv5 with oriented bounding boxes\cite{hukaixuan199706272021} is chosen for its simplicity and wide usage in practical settings.
To enable this lightweight model to do what humans do---exploit the temporal consistency of video frames and the information induced by motion---we add information on the last frame into the current one.
We encode this additional information in the existing RGB channels:

The \emph{red channel} is replaced by a grayscale representation of the original image.
Even though the gray-scaled image is harder to interpret even for humans, the general shape of the objects in the picture is still intact.
Therefore, the filters of the convolutional layers that are applied to this channel would still be able to detect shapes.
This decision was influenced by the lack of information in color concerning the object classes.

The \emph{green channel} is repurposed to represent large pixel changes in comparison to the previous frame.
Hence, the channel indicates movement to the model if detected to encourage learning to distinguish object classes that move more frequently (e.g. staff members) from those that do not (like the typically stationary patients).
This simplistic pixel change indication is only applicable because the video is observing a relatively still environment, and the camera does not change its angle, position, or perspective.

The \emph{blue channel} is replaced by a bitmap that contains the area of the previous frame's bounding boxes (either ground truth during training or predictions during inference), marked by an arbitrary value ($32$), and otherwise $0$.
This encourages the model to consider the bounding boxes from the predictions or the ground truth of the last frame for the current frame.
To prevent an overreliance on this channel, only a randomly chosen half of the samples have non-zero bitmaps in the third channel.
Of the half that has bounding box information from the previous frame, $20$\% were randomly chosen to be discarded completely to account for new objects appearing in the image, as well as for missing detections from previous frames. 
Furthermore, $60$\% of the bounding boxes' areas were randomly moved around up to $10$ pixels according to a uniform distribution to account for minor local variations in the earlier frame's predictions, just as object movement.

With this repurposing of the RGB channels, we provide the model with useful information about the temporal consistency of the frame succession (and thus object movement) without having to adapt YOLOv5's efficiently executable and well-proven architecture, thus making development convenient.
We hypothesize that this gain in temporal context information more than compensates for the loss of color information through the reduction of the current frame to grayscale, leading to higher OD rates in the experiments.

\section{Experimental Setup and Results}
\label{sec:Result}

\begin{figure}[t]
    \centering
    \includegraphics[width=.49\linewidth]{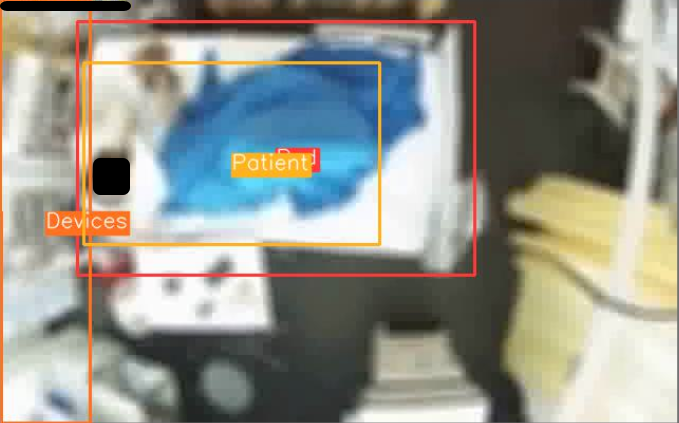} 
    \includegraphics[width=.49\linewidth]{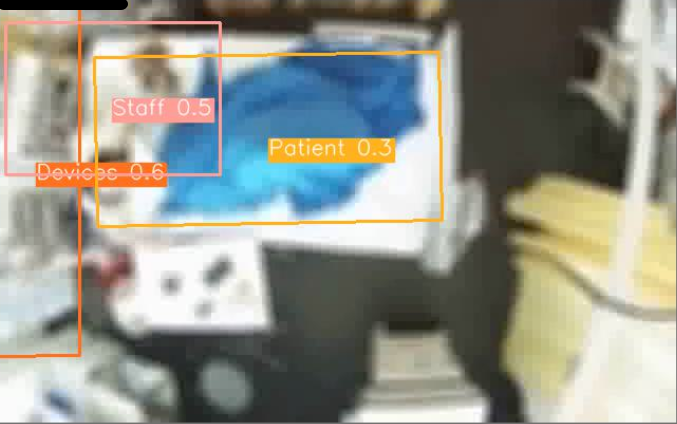} 
    \includegraphics[width=.49\linewidth]{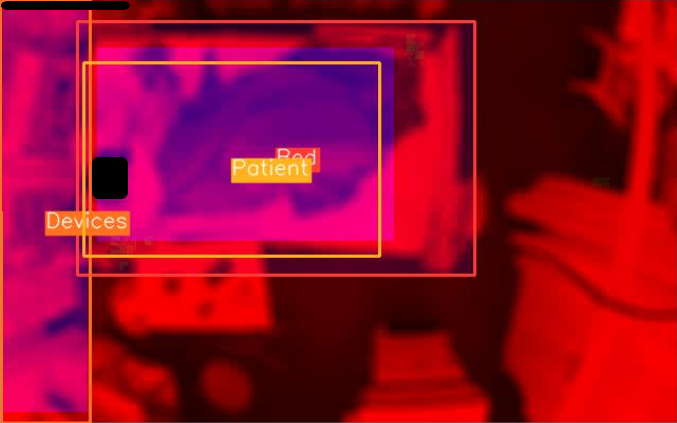} 
    \includegraphics[width=.49\linewidth]{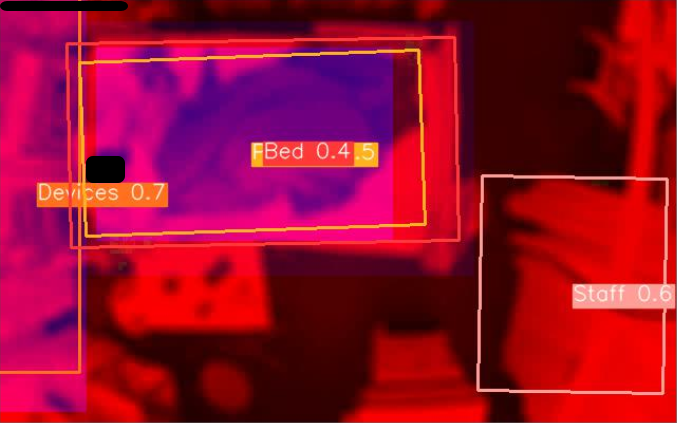} 
    \caption{
    OD ground truth (left) and the predictions (right) on a sample image.
    The top row shows the baseline model in- and output, and the bottom row the proposed model (grayscale image in red channel).
    }
    \label{fig:SampleOutputs}
\end{figure}

\noindent\textbf{Data Collection}
\label{sssec:ContextDataCollection}
For the development of the OD models, we prospectively collected blurred anonymized video data from cameras (AXIS M1065-L) directed onto the bedsides of a 12-bed neurocritical care unit at the University Hospital Zurich.
The blurred video streams have a resolution of $640 \times 400$ pixels at $25$ frames per second, collected by a dedicated research IT infrastructure \cite{Boss2022}.
The video data streams are blurred using a software solution for video stream conversion (FFmpeg, \url{https://ffmpeg.org/}) with a box blur filter ({\small\texttt{boxblur=6:1}}).
The blurring is required to ensure the privacy of clinical and hospital staff members as well as visitors of patients.
Written informed consent was received by all patients or by their legal representatives.
The study (part of the project ``ICU Cockpit'') was approved by the ethics committee of Kanton Zurich (Basec no. 2021-01089), Switzerland, and was conducted following the ethical standards of the 2013 declaration of Helsinki for research involving human subjects.

\begin{figure}[t]
    \centering
    \begin{tikzpicture}
        \begin{axis}[
                height=.6\linewidth, width=.8\linewidth,
                colormap={bluewhite}{color=(white) rgb255=(40,70,240)},
                xlabel=\scriptsize{}Ground Truth,
                xlabel style={yshift=0pt},
                ylabel=\scriptsize{}Predicted (baseline),
                ylabel style={yshift=0pt},
                yticklabel style={rotate=90},
                xticklabels={Bed, Staff, Devices, Patient, BG}, 
                xtick={0,...,4}, 
                xtick style={draw=none},
                yticklabels={Bed, Staff, Devices, Patient, BG}, 
                ytick={0,...,4}, 
                ytick style={draw=none},
                enlargelimits=false,
                colorbar,
                colorbar style={font=\scriptsize},
                xticklabel style={
                  rotate=0
                },
                yticklabel style={/pgf/number format/fixed, font=\tiny},
                xticklabel style={/pgf/number format/fixed, font=\tiny},
                nodes near coords={\pgfmathprintnumber\pgfplotspointmeta},
                nodes near coords style={font=\scriptsize,
                    yshift=-5pt, /pgf/number format/.cd, fixed, precision=2
                },
            ]
            \addplot[
                matrix plot,
                mesh/cols=5, 
                point meta=explicit,draw=gray
            ] table [meta=C] {
                x y C
                0 0 0.311
                1 0 0
                2 0 0
                3 0 0.02
                4 0 0.05
                
                0 1 0
                1 1 0.71
                2 1 0
                3 1 0
                4 1 0.83
                
                0 2 0
                1 2 0
                2 2 0.93
                3 2 0
                4 2 0.01
        
                0 3 0.04
                1 3 0
                2 3 0
                3 3 0.69
                4 3 0.11
        
                0 4 0.65
                1 4 0.29
                2 4 0.07
                3 4 0.3
                4 4 0
                
            }; 
        \end{axis}
    \end{tikzpicture}
    \begin{tikzpicture}
        \begin{axis}[
                height=.6\linewidth, width=.8\linewidth,
                colormap={bluewhite}{color=(white) rgb255=(40,70,240)},
                xlabel=\scriptsize{}Ground Truth,
                xlabel style={yshift=0pt},
                ylabel=\scriptsize{}Predicted (Ours),
                ylabel style={yshift=0pt},
                yticklabel style={rotate=90},
                xticklabels={Bed, Staff, Devices, Patient, BG}, 
                xtick={0,...,4}, 
                xtick style={draw=none},
                yticklabels={Bed, Staff, Devices, Patient, BG}, 
                ytick={0,...,4}, 
                ytick style={draw=none},
                enlargelimits=false,
                colorbar,
                colorbar style={font=\scriptsize},
                xticklabel style={rotate=0},
                yticklabel style={/pgf/number format/fixed, font=\tiny},
                xticklabel style={/pgf/number format/fixed, font=\tiny},
                nodes near coords={\pgfmathprintnumber\pgfplotspointmeta},
                nodes near coords style={font=\scriptsize,
                    yshift=-5pt, /pgf/number format/.cd, fixed, precision=2
                },
            ]
            \addplot[
                matrix plot,
                mesh/cols=5, 
                point meta=explicit,draw=gray
            ] table [meta=C] {
                x y C
                0 0 0.95
                1 0 0
                2 0 0
                3 0 0.08
                4 0 0.01
                
                0 1 0
                1 1 0.75
                2 1 0
                3 1 0
                4 1 0.89
                
                0 2 0
                1 2 0
                2 2 1.0
                3 2 0
                4 2 0.01
        
                0 3 0.02
                1 3 0
                2 3 0
                3 3 0.82
                4 3 0.09
        
                0 4 0.04
                1 4 0.25
                2 4 0
                3 4 0.10
                4 4 0
                
            }; 
        \end{axis}
    \end{tikzpicture}
    \caption{
    Confusion matrices for the baseline (top) and proposed model (bottom).
    Note that ``BG'' means ``Background''.
    }
    \label{fig:ConfusionMatrix}
\end{figure}

\begin{table}[t]
    \centering
    \begin{tabular}{l|c|llll|l}
    \multirow{2}{*}{Model}&\multirow{2}{*}{Epochs}&            \multicolumn{5}{c}{mAP@$.$5(\%)}            \\\cline{3-7}
                    &           & Bed              & Staff         & Devices       & Patient       & All   \\\hline
        YOLOv5      &       119 & 98.0             & \textbf{58.1} & 97.6          & 95.3          & 87.2  \\
        Proposed    &        10 & \textbf{99.5}    & \textbf{58.1} & \textbf{98.4} & \textbf{99.4} & \textbf{88.9}
    \end{tabular}
    \caption{
    Amount of received training (epochs) and mAP@.5 for the baseline and proposed model by class and overall.
    }
    \label{tab:ResultsMAP}
\end{table}

\noindent\textbf{Data Preprocessing and Labeling}
\label{sssec:ContextVideoPreprocessing}
Videos are collected as 24-hour recordings ensuring the capturing of different lighting conditions as well as situations from night and day shifts.
The 24-hour recordings are cut such that individual videos can be attributed to individual patients; only videos from patients who have given informed consent are kept.
To extract time periods that show movement automatically, pixel-wise differences between frames are calculated and used as a metric for general motion.
By choosing an appropriate threshold, we can identify video snippets that included medical personnel with a high probability.
An additional $10$~s of video data is added to either side of the identified video snippets, resulting in videos that typically range from one to several minutes in length.
Through this process, in total $30.748$ clips are accumulated.
From these clips, $196$ clips are selected to be hand-labeled, with a balanced representation of the different beds and scenarios.
The chosen labels are ``patient'', ``bed'', ``staff'', and the location of the (medical) ``devices''---as the corresponding objects are the ones that play the most crucial role in establishing context for the situation at a patient’s bedside.
The bounding boxes of the labels can be rotated as well; e.g. the bed frame and the patient laying within, or staff walking around, turning, or leaning over the bed and patient.

\noindent\textbf{Clinical Compute Infrastructure}
\label{sssec:ContextConstraintsInfrastructure}
Due to the handling of patient-related data, hardware options for model training are severely restricted.
The only compliant option is a virtual machine with $8$ CPUs ($2$ GHz), $32$ GB RAM, $500$ GB SSD storage, and two NVIDIA TITAN V GPUs.
For deployment, the final OD model would run on a server without GPUs.

\noindent\textbf{Baseline and Training Details}
\label{ssec:MethodBaseline}
The YOLOv5 ``s''-version is chosen for both the baseline and proposed method as it is the second smallest one and therefore very attractive to be deployed into the production environment without GPUs.
To train the models, the dataset is split on a frame-by-frame basis into $70$\% training data, $15$\% validation data, and $15$\% test data.
Both models were trained until they did not show any improvement for $100$ consecutive epochs up to maximum $300$ epochs.
They are trained using the default hyperparameters provided by YOLOv5 ($0.01$ learning rate and $0.937$ momentum after a warm-up period of three epochs; before that, $0.1$ learning rate and $0.8$ momentum).
The model used an L2 regularization (weight decay) of $0.0005$.

\noindent\textbf{Results}
Models are evaluated using the mean average precision (mAP) metric at $0.5$ IoU overlap as well as the amount of training measured in epochs.
As shown in \Cref{tab:ResultsMAP}, the training of the baseline model lasted for $119$ epochs (after which it showed no further improvement), while the proposed method required only a fraction ($8.4$\%) of this training time.
As seen in both the confusion matrix and the example outputs in \Cref{fig:ConfusionMatrix,fig:SampleOutputs} respectively, the baseline model has trouble recognizing members of staff and the bed frame.
While our proposed model also struggles with the staff category, it does significantly better on bed frames and patients.
When comparing the mean average precision at $0.5$ IoU overlap, the proposed method outperforms the baseline by $1.7$\% (refer to \Cref{tab:ResultsMAP}).

\section{Discussion and Conclusions}

We demonstrated that our proposed method outperforms the baseline model using only a fraction of the training time.
We attribute this improvement to the new data format in which information is presented to the model in the repurposed image channels, enabling it to learn from temporal correlations.
While it loses the color information, the additional information about pixel changes and bounding boxes from the earlier frame more than compensates for this loss, evidenced by the performance increase and shortened training time.
What seems like a hack is typical for deep learning in practice:
In absence of large training sets and conditions as found in public benchmarks \cite{stadelmann2019beyond}, the available information has to be exploited optimally while considering computational boundary conditions.

We identify great potential in exploring the proposed method further, including not replacing the RGB channels, but instead expanding them with additional channels.
There is also the question of the efficacy of this method when not using stationary videos, but using changes in perspective, viewing angle, etc.
However, we hypothesize that it will increase the efficiency of training of any given model, and leave respective experiments to future work, together with necessary ablation studies w.r.t. hyperparameters that did not fit the scope of this short communication.
Note that the proposed method is model-independent and applicable to architectures beyond YOLOv5.

From a medical AI perspective, we anticipate that even the limited contextual information extracted by the proposed method can contribute significantly to improved artifact detection and handling.
For example, staff presence can now be used during a preprocessing step or directly as input to other machine-learning models to reduce false alarms due to wrong measurements.
A different application would be to determine the level of care received by individual patients to optimize the assignment of nursing staff and anticipate possible situations of nursing overload.

\noindent\textit{Acknowledgement} This work has been supported by DIZH grant ``AUTODIDACT'', the SNSF, and the CSB Berlin.

\bibliographystyle{ieeetr}
\bibliography{references_boss,references_embe}

\end{document}